\documentclass{article}

\usepackage{arxiv}

\usepackage{graphicx}
\usepackage[utf8]{inputenc} % allow utf-8 input
\usepackage[T1]{fontenc}    % use 8-bit T1 fonts
\usepackage[colorlinks,allcolors=blue]{hyperref}       % hyperlinks
\usepackage{url}            % simple URL typesetting
\usepackage{booktabs}       % professional-quality tables
\usepackage{amsfonts}       % blackboard math symbols
\usepackage{nicefrac}       % compact symbols for 1/2, etc.
\usepackage{microtype}      % microtypography
\usepackage{lipsum}		% Can be removed after putting your text content
\usepackage{graphicx}
\usepackage{natbib}
\usepackage{doi}
\usepackage{tabularx,booktabs}
\usepackage{algorithm}
\usepackage{algpseudocode} %not in acm list
\usepackage{multirow}
\usepackage{xcolor}
\usepackage{multicol}
\usepackage{subcaption}
\usepackage{tikz} % not in acm list
\usepackage{amsmath}
\usepackage{textcomp}
\usepackage{libertine}
\usepackage{comment}
\usepackage{amssymb}
\algnewcommand\algorithmicforeach{\textbf{for each}}
\algnewcommand\algorithmicdoparallel{\textbf{do in parallel}}
\algdef{S}[FOR]{ForEach}[1]{\algorithmicforeach\ #1\ \algorithmicdo}
\algdef{S}[FOR]{ForEachParallel}[1]{\algorithmicforeach\ #1\ \algorithmicdoparallel}
\algnewcommand{\sIf}[2]{%\sIf{<if>}{<then>}
  \State \algorithmicif\ #1\ \algorithmicthen\ #2}

\graphicspath{graphics}
\usepackage{xspace}

\title{Comparing the Digital Annealer with Classical Evolutionary Algorithm}

\author{ \href{https://orcid.org/0000-0003-0854-4777}{\includegraphics[scale=0.06]{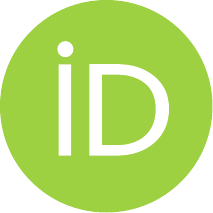}\hspace{1mm}Mayowa Ayodele}\\
	Fujitsu Research of Europe Ltd.\\
	The Urban Building, 3-9 Albert Street\\
	Slough, United Kingdom,  SL1 2BE\\
	\texttt{mayowa.ayodele@fujitsu.com} \\
}

\renewcommand{\shorttitle}{Comparing the Digital Annealer with Classical Evolutionary Algorithm}
\begin{document}
\maketitle

\begin{abstract}
 In more recent years, there has been increasing research interest in exploiting the use of application specific hardware for solving optimisation problems. Examples of solvers that use specialised hardware are IBM's Quantum System One and D-wave's Quantum Annealer (QA) and Fujitsu's Digital Annealer (DA). These solvers have been developed to optimise problems faster than traditional meta-heuristics implemented on general purpose machines. Previous research has shown that these solvers (can optimise many problems much quicker than exact solvers such as GUROBI and CPLEX. Such conclusions have not been made when comparing hardware solvers with classical evolutionary algorithms. 

Making a fair comparison between traditional evolutionary algorithms, such as Genetic Algorithm (GA), and the DA (or other similar solvers) is challenging because the later benefits from the use of application specific hardware while evolutionary algorithms are often implemented on general-purpose machines. Moreover, quantum or quantum-inspired solvers are limited to solving problems in a specific format. A common formulation used is Quadratic Unconstrained Binary Optimisation (QUBO). Many optimisation problems are however constrained and have natural representations that are non-binary. Converting such problems to QUBO can lead to more problem difficulty and/or larger search space.

The question addressed in this paper is whether quantum or quantum-inspired solvers can optimise QUBO transformations of combinatorial optimisation problems faster than classical evolutionary algorithms applied to the same problems in their natural representations. We show that the DA often present better average objective function values than GA on Travelling Salesman, Quadratic Assignment and Multi-dimensional Knapsack Problem instances.
\end{abstract}



\section{Introduction}
Optimisation problem solvers that use specialised hardware, particularly those classified as quantum or quantum-inspired solvers, have been of interest in more recent years. These solvers are however limited by the size of problems they can solve. D-wave's Advantage released in 2020, which uses quantum hardware, has the capability of about 5,000 quantum bits (qubits) \cite{mcgeoch2020d} while IBM's Eagle has announced extending its 27-qubit solver to 127-qubit \cite{chow2021ibm}. Quantum-inspired solvers can solve larger problems e.g. the Digital Annealer (DA) which uses application specific CMOS and is able to solve binary quadratic problems with up to 100,000 bits \cite{da3}. %A detailed analysis of hardware solvers within the context of Ising machines is presented in \cite{mohseni2022ising}.

Quantum and quantum-inspired optimisation solvers are applied to specific formulations. Quadratic Unconstrained Binary Optimisation (QUBO) is a common formulation used by these solvers. Combinatorial optimisation problems are therefore initially converted to QUBO before these solvers can be applied to them. QUBO problems are unconstrained, quadratic and of binary form generally defined as follows:
\begin{equation}
\label{eq:qubo}
  E(x) = x^TQx + q \;,
\end{equation}
where $Q$ represents a $m \times m$ matrix, $q$ is a constant term, a solution $x=(x_1, \dots,x_m)$ is an $m$-dimensional vector, and $E(x)$ is the \emph{energy} (or fitness) of $x$. 

Quantum and quantum-inspired optimisation methods are largely, but not solely, inspired by physics e.g quantum annealing and Simulated Annealing (SA) \cite{aramon2019physics}. However, inspiration of evolutionary algorithm originally came from biology. The concept of natural selection and the principle of ``survival of the fittest" form the core of evolutionary algorithms. Genetic Algorithms (GA) are some of the most widely studied algorithms in evolutionary computation and dates back to the 1960s \cite{katoch2021review}. GAs use three main concepts; selection, crossover and mutation. Another evolutionary algorithm that has been well studied is Differential Evolution (DE). DE, like the GA, uses crossovers and mutation operators \cite{storn1997differential}. Estimation of distribution Algorithms (EDAs) follow similar idea as the GA, however probabilistic models are used to generate new solutions \cite{larranaga2001estimation}. 

An optimisation problem that has been widely studied in operations research is the 0-1 knapsack problem \cite{salkin1975knapsack}, Multi-dimensional Knapsack Problem (MKP) is one of its variants. MKP is particularly interesting because it consists of multiple inequality constraints. Algorithms such as GA \cite{chu1998genetic} and EDA \cite{wang2012effective} have been applied to the MKP. The DA has also been applied to the MKP in \cite{Diez2022Exact}.

Permutation problems are also well-studied combinatorial optimisation problems in nature-inspired computing. They have many real-world applications especially in planning and logistics. Some of the most frequently studied permutation problems in literature are the well-known Travelling Salesman Problem (TSP) and Quadratic Assignment Problem (QAP). Since these problems are NP-hard, heuristics and meta-heuristics have been proposed for solving them. Many evolutionary algorithms have been applied to the TSP or QAP e.g. EDA \cite{ayodele2016rk}, GA \cite{grefenstette1985genetic,tate1995genetic,ahuja2000greedy} and Differential Evolution \cite{mi2010improved,hameed2018improved}. These problems have also gained popularity within the context of solving optimisation problems with quantum or quantum-inspired algorithms e.g DA \cite{matsubara2020digital,ayodele2022penalty} and Quantum Annealer (QA)\cite{warren2020solving,kuramata2021larger}. While some algorithms are designed to solve permutation problems in their natural representation, others require alternative representations. Some examples of alternative representations used in evolutionary algorithms are random keys \cite{ayodele2016rk}, factoradics \cite{regnier2014factoradic}, binary \cite{baluja1994population} and matrix representation \cite{larranaga1999genetic}. When formulation permutation problems as QUBO such that quantum or quantum-inspired algorithms can be used to solve them, the two-way one-hot representation also known as permutation matrix is often used \cite{lucas2014ising,matsubara2020digital,ayodele2022penalty}.

In \cite{khumalo2021investigation}, two optimisation methods; Branch and Bound and SA implemented on classical computers were compared to the Variational Quantum Eigensolver (VQE) algorithm and Quantum Approximate Optimisation Algorithm (QAOA) implemented on IBM’s suite of Noisy Intermediate Scale Quantum devices. Comparison was made using QAP and TSP instances. The study concluded that better performance was achieved using the classical methods. This is not unexpected because converting simple permutation problems to QUBOs significantly increases the search space. The search space also consists of more infeasible than feasible solutions.

In this study, we compare GA with the DA on MKP, QAP and TSP instances. GA was chosen because it is the most studied evolutionary algorithm in literature. DA was also chosen because it is able to handle larger problem sizes compared to other similar solvers e.g. QA. GA implementation presented in \cite{pymoo} and third generation DA \cite{da3} are used in this work. MKP, QAP and TSP were particularly chosen because of their complexity from the perspective of solving the resulting QUBO. 

Comparing the GA with DA fairly is challenging for the following reasons. 
\begin{itemize}
    \item  Problem formulations: GA is not restricted to only one formulation like the DA, giving the GA an advantage over the DA.
    \item Hardware differences: DA has hardware advantage as it uses application specific hardware designed to run faster than general purpose machines often used to implement GAs. This therefore gives the DA an advantage over the GA.
    \item Hyperparameter tuning: the parameters of the GA used in this study needs to be tuned by the user. DA however has inbuilt parameter tuning feature. 
    \item Performance metric: while the number of solution evaluations is a common metric in evolutionary computation, time to solution is often used when comparing hardware solvers such as DA. DA neither keep a history of the number of solutions visited nor do complete evaluations, it estimates the difference in energy between a solution and all its neighbours in constant time. The classic GA however does complete solution evaluations.
\end{itemize}

Although these factors will affect the result of this research, comparing the DA with GA is useful from a user perspective where the aim is to get a solution to a problem as quickly as possible regardless of how the solution is derived. It is also useful for the research community to understand the extent to which one might be better than the other and on which problem categories. 

To address or minimise the effect of the factors that limit a fair comparison between the DA and GA, we do the following.
\begin{itemize}
 \item  Problem formulations: the DA is allowed to solve the considered problems in binary and quadratic form while the GA is allowed to solve problems in their natural formulation. This is because both algorithms have been optimised to solve problems in this form. 
\item Hardware differences: to minimise the hardware advantage, experiments carried out using the GA benefited from ``Just in Time'' (JIT) compilation with numba python library, which can achieve a speed up of up to approximately 500 times \cite{lam2015numba}. Furthermore, we explored both serialised and parallelised evaluations. Parallel evaluation is done using \textit{Threaded Loop-wise Evaluation} described in \cite{pymoo}. High performance machine equipped with Ubuntu 18.04, Intel Xeon Gold 5218 CPU @ 2.30GHz, and 192GB Memory was also used when running the GA. 
 \item Hyperparameter tuning: to allow for a fair comparison, we used the same hyper-parameter tool ``Optuna'' \cite{optuna_2019} for both DA and GA.
 \item Performance metric: since the GA and DA can be terminated after a specified time limit, we set the same time limit for both algorithms. The aim is to address the question of how much quicker a solution can be attained using a hardware solver like the DA compared to a classic GA implemented on a general-purpose computer, or vice versa.
\end{itemize}

The rest of this paper is structured as follows. The problem formulations are presented in Section \ref{sec:pe}. The Algorithms used in this study are described in \ref{sec:algorithms}. Section \ref{sec:ex} describes the problem sets and the parameter settings. Results are analysed in section \ref{sec:res}. Finally, conclusions and further work are presented in Section \ref{sec:con}.

\section{Problem Definitions}
\label{sec:pe}
This sections defines the formulations of the MKP, QAP and TSP used in this study.

\subsection{Binary Problem}
Unlike other problem classes that would need to be converted from their natural representations to binary, binary problems are able to retain their representation when converted to QUBO. The MKP problem used in this study are described as follows.

\subsubsection{Multi-dimensional 0-1 Knapsack Problem}
0-1 knapsack problems are naturally represented with binary variables. The MKP consists of $n$ items and $m$ constraints. Each item $i$ is defined by a value $p_i$ and units of resource consumption for each constraint $k$, $w_{ik}$. The aim is to find a subset of items that lead to maximum profit without exceeding any of the capacities $W = \left \{ W_1,\cdots, W_m\right \}$. We convert this to a minimisation problem by converting the profits to negative values.

\begin{align}
\label{eq:mknpcs}
    \text{minimise}    \quad & f(x) = -\sum_{i=1}^{n} p_i {\,} x_i  \enspace , \\
    \label{eq:mknpcs2}
    \text{subject to} \quad & \left ( \sum_{i=1}^{n} w_{ik} \cdot x_i \right ) \leq W_k, \quad x_i \in \{0, 1\} \enspace .
\end{align}

In the QUBO formulation, cost function $c(x)$ of the knapsack problem is the same as $f(x)$. To represent the constraint function $g(x)$, an approach similar to one used in QUBO formulation of 0-1 knapsack problem in \cite{coffey2017adiabatic} is used. The constraint function using slack variables is presented in Eqs.\eqref{eq:mk2}.

\begin{equation}
\label{eq:mk2}
   g\left ( x \right ) = \sum_{k=0}^{m} \left ( \sum_{j=0}^{M_k-1}2^jy_{j} + \left (W_k + 1 -2^{M_{k}}\right )y_{M_k} - \sum_{i=1}^{n}w_{ik}x_i \right)^2\\
\end{equation}

In Eq.~\eqref{eq:mk2}, $2^{M_k} \leq W_k < 2^{M_{k} + 1}$ where $M_k = \log_{2} W_k$. Slack variables $y$ are represented in powers of two. So the number of slack variables used for each capacity constraint is limited to $\log_{2} W_k + 1$ for the slack range $\{0, \ldots, W_k\}$. The total number of binary variables used to represent this problem is therefore $n + \sum_{k=0}^{m} \left(\log_{2} W_k + 1\right)$. 

DA can be applied to either the QUBO representing $g(x)$ (Eq. \eqref{eq:mk2}) or the inequality constraints (Eq. \eqref{eq:mknpcs2}). GA is applied to the standard formulation in Eq. \eqref{eq:mknpcs} and \eqref{eq:mknpcs2}.

\subsection{Permutation Problems}
A valid permutations is described as $\pi = \left \{ \pi_1,\ldots, \pi_n \right \}$, where  $\pi_i \neq \pi_j\ \forall\ i \neq j$. Representations of permutation problems are presented as follows.

\begin{itemize}
    \item{\textbf{Permutation:}}
    permutation representation is often used to describe the order in which a set of activities are carried out. Each unique value in the permutation represents some activity while the position of each value is the rank of the corresponding activity. Activities are performed in order of their ranks. An example is shown as follows:
    
     \begin{equation}
        \label{eq:permutation}
    \begin{bmatrix}
    2 & 3  & 1
    \end{bmatrix}\ \Leftrightarrow \ activity2\rightarrow activity3\rightarrow activity1. \nonumber
     \end{equation}

    \item {\textbf{Two-way one-hot:}}
    in the two-way one-hot representation, each unique value in the permutation is represented by a vector of zeros where only one bit representing such value is set to 1. An example is shown as follows.
    
      \begin{equation}
        \label{eq:2w1h2}
        \begin{bmatrix}
            2 &  3& 1
            \end{bmatrix} 
             \Leftrightarrow 
         \begin{bmatrix}
            0 & 1 & 0\\ 
            0 & 0 & 1\\ 
            1 & 0 & 0
            \end{bmatrix} 
    \end{equation}

    \begin{equation}
        \label{eq:2w1h1}
         \begin{bmatrix}
            0 & 1 & 0\\ 
            0 & 0 & 1\\ 
            1 & 0 & 0
            \end{bmatrix} 
             \Leftrightarrow 
        \begin{bmatrix}
            0 & 1 & 0 & 0 & 0 & 1 & 1 &  0& 0
            \end{bmatrix}
    \end{equation}

    For a solution to therefore be valid, i.e. can be decoded to a valid permutation, each row and column of the two-way one-hot solution shown in Eq. \eqref{eq:2w1h2} (right) must sum to one. In the QUBO formulation, we achieve this by penalising any solution that cannot be decoded to a valid permutation. The constraint function is shown in Eq.~(\ref{eq:penfun}), $g(x) = 0$ in feasible solutions and $g(x) > 0$ in infeasible solutions.
    
    \begin{equation}
    \label{eq:penfun}
    g(x) = \sum_{i=1}^{n}\left ( 1- \sum_{k=1}^{n} x_{i,k}\right )^2 + \sum_{k=1}^{n}\left ( 1- \sum_{i=1}^{n} x_{i,k}\right )^2 \;.
    \end{equation}

\end{itemize}

\subsubsection{Quadratic Assignment Problem}
    The QAP can be described as the problem of assigning a set of $n$ facilities to a set of $n$ locations. For each pair of locations, a distance is specified. For each pair of facilities, a flow (or weight) is specified. The aim is to assign each facility to a unique location such that the sum of the products between flows and distances is minimised. 
    
    Formally, the QAP consists of two $n \times n$ input matrices $H=[h_{ij}]$ and $D=[d_{kl}]$, where $h_{ij}$ is the flow between facilities $i$ and $j$, and $d_{kl}$ is the distance between locations $k$ and $l$, the solution to the QAP is a permutation $\pi = (\pi_1, \ldots ,\pi_n )$ where $\pi_i$ represents the location that facility $i$ is assigned to. The objective function (total cost) is formally defined as follows
    
    \begin{equation}
        \label{eq:fqap}
        \text{minimise}\ f(\pi) = \sum_{i=1}^{n}\sum_{j=1}^{n}h_{ij} \cdot d_{\pi_i, \pi_j}\;.
    \end{equation}
    
    The cost function, $c(x)$ of the QUBO representing the QAP is presented in Eq.~(\ref{eq:cqap}) and the constraint function $g(x)$ is the same as Eq.~(\ref{eq:penfun}).
    
    \begin{equation}
    \label{eq:cqap}
      c\left ( x \right ) = \sum_{i=1}^{n}\sum_{j=1}^{n}\sum_{k=1}^{n}\sum_{l=1}^{n}h_{ij}d_{kl}x_{ik}x_{jl}\;.
    \end{equation}

\subsubsection{Travelling Salesman Problem}
    The TSP consists of $n$ locations and a matrix $d$ representing distances between any two locations. The aim of the TSP is to minimise the distance travelled while visiting each location exactly once and returning to the location of origin. Given that $\pi_i$ is used to denote the $i^{th}$ city and $d_{\pi_{i-1},\pi_i}$ is the distance between $\pi_i$ and $\pi_{i-1}$. The solution to the TSP is a permutation $\pi = \left \{ \pi_1,\ldots ,\pi_n \right \}$ where each $\pi_i\ (i= 1,\ldots ,n)$ represents the $i^{th}$ location to visit. The TSP is formally defined as

    \begin{equation}
    \label{eq:tsp}
        \text{minimise } f\left ( \pi \right ) = \sum_{i=2}^{n} d_{\pi_{i-1}, \pi_i} + d_{\pi_n, \pi_1}\;.
    \end{equation}

    The cost function, $c(x)$ of the QUBO representing the TSP is presented in Eq.~(\ref{eq:qtsp}) and the constraint function $g(x)$ is the same as Eq.~(\ref{eq:penfun}).

    \begin{equation}
    \label{eq:qtsp}
    c\left ( x \right ) = \sum_{(l,i) \in E}d_{l,i} \sum_{k=1}^{n}x_{l,k}x_{i,k+1} \;.
    \end{equation}

For both TSP and QAP formulations, $c\left ( x \right ) \equiv  f\left ( \pi \right )$. When we apply the GA, we minimise $f\left ( \pi \right )$ (Eq. \eqref{eq:fqap} for QAP or Eq. \eqref{eq:tsp} for TSP). However, when the DA is applied, we solve two QUBOs $C$ (representing $c\left ( x \right )$) and $G$ (representing $g\left ( x \right )$). The DA applies its algorithm to a weighted aggregate of $C$ and $G$ (i.e. $Q$ = $C + \alpha \cdot G$). Note that $\alpha$ is the penalty weight applied to the constraint function.

\section{Algorithms}
\label{sec:algorithms}
This section described the algorithms used in this study; DA and GA.
\subsection{Digital Annealer}
\label{sec:da}
The algorithm that supports the $1^{st}$ generation DA \cite{aramon2019physics} is presented in Alg. \ref{alg:da}.

\begin{algorithm}
\caption{The DA ($1^{st}$ generation) Algorithm}
\label{alg:da}
\begin{algorithmic}[1]
\State initial\_state $\leftarrow$ an arbitrary state \label{a1}
\ForEach {run}
\State initialise to initial\_state
\State $E_{\text{offset}}\ \leftarrow\ 0$
\ForEach {iteration }
\State update the temperature if temperature update due 
\ForEach {variable $j$, in parallel }
\State propose a flip using $\Delta E_j - E_{\text{offset}}$
\State if acceptance criteria ($P_j$) is satisfied, record \label{a2}
\EndFor
\If{at least one flip is recorded} 
    \State chose one flip at random from recorded flips
    \State update the state and effective fields, in parallel
    \State $E_{\text{offset}}\ \leftarrow\ 0$
\Else
    \State $E_{\text{offset}} = E_{\text{offset}} +$  offset\_increase\_rate
\EndIf
\EndFor
\EndFor
\end{algorithmic}
\end{algorithm}

The first-generation DA is a single flip solver with similar properties as the Simulated Annealing (SA). It is however designed to be more effective than the classical SA algorithm \cite{aramon2019physics}. The DA can evaluate neighbouring solutions in parallel and in constant time regardless of the number of neighbours. DA does not completely evaluate each solution but computes the energy difference resulting from flipping any single bit of the parent solution. Furthermore, considering all neighbouring solutions rather than just one neighbour significantly improves acceptance probabilities compared to the regular SA algorithm. Also, the DA uses an escape mechanism to avoid being trapped in local optimal. As shown in the algorithm, $E_{\text{offset}}$ is used to relax acceptance criteria. The $E_{\text{offset}}$ is incremented by a parameter ($\text{offset\_increase\_rate}$) each time no neighbour which satisfies the acceptance criteria is found. The acceptance criteria is defined by $P_j = \text{exp}(\text{min}(0,-(\Delta E_j - E_{\text{offset}}) /\delta))$ where $P_j$ is the probability of accepting the $j^{th}$ flip. Note that $\Delta E_j$ represents the difference in energy as a result of flipping the $j^{th}$ bit and $\delta$ is the current temperature.

More extensive details of the algorithm can be seen in \cite{aramon2019physics}. The $1^{st}$ and $2^{nd}$ generation DAs were released in May 2018 and December 2018. Both versions were designed to solve optimisation problems that have been formulated as QUBOs. The most recent generation of the DA is the $3^{rd}$ generation which can find optimal or sub optimal solutions to Binary Quadratic Problems (BQP) of up to 100,000 bits \cite{da3}. BQPs include QUBO but also other binary and quadratic formulations that may have constraints. Note that the algorithm that supports the $3^{rd}$ generation DA has been updated to perform better than Alg. \ref{alg:da}. Some of the changes made to the $3^{rd}$ generation DA includes better support for two-way one-hot representation, inequality constraints and penalty weight settings $\alpha$ \cite{da3}\footnote{available from \url{https://www.fujitsu.com/jp/documents/digitalannealer/researcharticles/DA_WP_EN_20210922.pdf}}.

\subsection{Genetic Algorithm}

In this study, we use a simple GA implementation from Pymoo \cite{pymoo}.\footnote{available in \url{https://pymoo.org/algorithms/soo/ga.html\#nb-ga}} The GA algorithm is presented in Alg. \ref{alg:ga}. We initialise the GA with problem size $n$, population size $p$, selection type $s\-type$, crossover type $c\-type$, mutation type $m\-type$, crossover rate $a$, mutation rate $b$ and fitness function $f$. At each iteration, we perform selection, crossover, and mutation until stopping criteria is met. $P$ and $P_{new}$ in Lines \ref{alglnp1} - \ref{alglnp3} are respectively used to denote the parent and offspring populations. The offspring population $P_{new}$ are considered as the parent population $P$ in the next generation. In the GA implementation used, there is the option to eliminate of keep duplicate solutions in $P$. The best solution $y$ found so far is returned when the stopping criteria is met. The stopping criteria used in the work is time limit in seconds.

  \begin{algorithm}
    \caption{GA}
    \label{alg:ga}
    \begin{algorithmic}[1]
    \State initialise $n$, $p$, $s\-type$, $c\-type$, $m\-type$, $a$, $b$, $f$ \label{gainit}
    \State generate population $P$ of size $p$, where $|x| = n\ \forall\ x \in P$
    \While{Stopping criteria not met}
    \State calculate $f(x)\ \forall\ x \in P$
    \State update best solution $y,\ y \in P$
    \Repeat
    \State perform selection $s\-type$
    \State perform crossover $c\-type$ to generate $x$ with rate $a$
    \State perform mutation $m\-type$ on $x$ with rate $b$
    \State Add x to $P_{new}$ \label{alglnp1}
    \Until {$|P_{new}| = p$} \label{alglnp2}
    \State update $P$ with $P_{new}$ \label{alglnp3}
    \EndWhile
    \State \textbf{Return} $y$
    \end{algorithmic}
    \end{algorithm}

    A range of crossover and mutation types were explored in this study. Methods suited for binary problems were used for MKP while methods suited for permutation problems were used for QAP and TSP. The list of parameters used in this work are shown in Table \ref{tb:gaparams}. More detailed explanations about the genetic operators for the GA are presented in \cite{pymoo}.

\section{Experimental Setup}
\label{sec:ex}
    Problem sets and parameter settings used in this study are presented in this section.
    
    \subsection{Datasets}
    Common MKP, QAP and TSP instances are taken from ORLIB\footnote{\url{http://people.brunel.ac.uk/~mastjjb/jeb/orlib/files/mknap2.txt}}, QAPLIB \cite{burkard1997qaplib} and TSPLIB \cite{reinelt1991tsplib} respectively. Tables \ref{tb:instmkp}, \ref{tb:instqap} and \ref{tb:insttsp} show the original problem size $n$ as well as the QUBO size $m$ for the MKP, QAP and TSP instances respectively. In the MKP, the DA can handle inequality constraints. If this approach is used, the size of the QUBO representing the objective function is the same as the original problem size $n$. However, where slack variables are used, the QUBO size is shown in column ``$m$ (slack)''. For the TSP, the first city was fixed to 1, we therefore only search for the ordering of entities 2 to total number of cities. The problem size $n$ was therefore reduced by 1. The QUBO size of the TSP and QAP is $m = n^2$.

     \begin{table}[t]
    \centering
    \caption{MKP instances and their solution sizes \label{tb:instmkp}}
    \resizebox{0.7\columnwidth}{!}{
    \begin{tabular}{@{}ccc@{}}
    \toprule
    \begin{tabular}[c]{@{}c@{}}MKP \\ Instances \end{tabular}& \begin{tabular}[c]{@{}c@{}}$n $ or\\ $m$ (ineq) \end{tabular} & \begin{tabular}[c]{@{}c@{}} $m$ \\ (slack)\end{tabular}\\
    \midrule
    weing1  & 28 & 50 \\
    weing2  & 28 & 50 \\
    weing3  & 28 & 50 \\
    weing4  & 28 & 50 \\
     \bottomrule
    \end{tabular}
    \begin{tabular}{@{}ccc@{}}
    \toprule
    \begin{tabular}[c]{@{}c@{}}MKP \\ Instances \end{tabular}& \begin{tabular}[c]{@{}c@{}}$n $ or\\ $m$ (ineq) \end{tabular} & \begin{tabular}[c]{@{}c@{}} $m$ \\ (slack)\end{tabular}\\
    \midrule
    weing5  & 28 & 50 \\
    weing6  & 28 & 50 \\
    weing7  & 105 & 131 \\
    weing8  & 105 & 131 \\ \bottomrule
    \end{tabular}}
    \end{table}

    \begin{table}[t]
    \centering
    \caption{QAP instances and their solution sizes \label{tb:instqap}}
    \resizebox{0.6\columnwidth}{!}{
    \begin{tabular}{@{}ccc@{}}
    \toprule
    QAP Instances & $n$ & $m$ \\
    \midrule
    had12 & 12 & 144 \\
    had14 & 14 & 196 \\
    had16 & 16 & 256 \\
    had18 & 18 & 324 \\
    had20 & 20 & 400 \\
     \bottomrule
    \end{tabular}
    \begin{tabular}{@{}ccc@{}}
    \toprule
    QAP Instances &$n$ &$m$ \\
    \midrule
    rou12 & 12 & 144 \\
    rou15 & 15 & 225 \\
    rou20 & 20 & 400 \\
    tai40a & 40 & 1600 \\
    tai40b & 40 & 1600 \\  \bottomrule
    \end{tabular}}
    \end{table}

     \begin{table}[t]
    \centering
    \caption{TSP instances and their solution sizes \label{tb:insttsp}}
    \resizebox{0.6\columnwidth}{!}{
    \begin{tabular}{@{}ccc@{}}
    \toprule
    TSP Instances &$n$ &$m$ \\
    \midrule
    bays29 & 28 & 784 \\
    bayg29 & 28 & 784 \\
    berlin52 & 51 & 2601 \\
    brazil58 & 57 & 3249 \\
    dantzig42 & 41 & 1681 \\
     \bottomrule
    \end{tabular}
    \begin{tabular}{@{}ccc@{}}
    \toprule
    TSP Instances &$n$ &$m$ \\
    \midrule
    fri26 & 25 & 625 \\
    gr17 & 16 & 256 \\
    gr21 & 20 & 400 \\
    gr24 & 23 & 529 \\
    st70 & 69 & 4761 \\ \bottomrule
    \end{tabular}}
    \end{table}

    QUBO matrices (in upper triangular format) representing the cost and constraint functions for QAP and TSP\footnote{\url{https://github.com/mayoayodelefujitsu/QUBOs}} as well as MKP\footnote{\url{https://github.com/mayoayodelefujitsu/MKP}} instances used are made available.

    \subsection{Parameter Settings}
     To find good sets of parameters for the GA and DA, a hyperparameter framework, Optuna \cite{optuna_2019} \footnote{available at \url{https://github.com/optuna/optuna}} was used. Optuna uses deep learning techniques to find hyper-parameters and has been used for parameter tuning in this study. Optuna has been used to search a range of values efficiently such that the algorithms can be executed with sets of parameters that work well on each individual problem. Full range of possible values have been used where reasonable; some values have been narrowed down to ranges that have been reported in previous work to lead to more promising performance. Optuna has been allowed to do 30 trials, each trial is assessed by the average fitness (20 runs) reached within a specified time limit. In this study we used 1 second time limit.
     
    \subsubsection{Genetic Algorithm}
    Ranges of parameters explored by Optuna for the GA are presented in Table \ref{tb:gaparams}. 
  
    \begin{table}[th]
    \begin{center}
    \caption{GA Parameters}
    \label{tb:gaparams}
    \resizebox{0.7\columnwidth}{!}{
    \begin{tabular}{@{}ccc@{}}
    \toprule
    \multirow{2}{*}{\textbf{Parameters}} &  \multicolumn{2}{c}{\textbf{Values}} \\ \cmidrule(l){2-3} 
     & Permutation & Binary \\ \midrule
    Population Size $p$ & $[n, 10n]$ & $[n, 10n]$ \\ \midrule
    Selection Type $s\-type$ & Random & Random\\ \midrule
    Crossover Type $c\-type$ & \begin{tabular}[c]{@{}c@{}}Order,\\ Edge Recombination\end{tabular} & \begin{tabular}[c]{@{}c@{}}One point, Two point,\\ Uniform, Exponential \end{tabular}  \\ \midrule
    Crossover Rate $a$  & {[}0.5, 0.9{]} & {[}0.5, 0.9{]} \\ \midrule
    Mutation Type $m\-type$  & Inverse  & Bit flip\\ \midrule
    Mutation Probability $b$ & {[}0, 0.2{]}  & {[}0, 0.2{]}\\ \midrule
    Eliminate Duplicates $dup$ & True, False  & True, False \\ \midrule
    Number of threads (pGA only) & 32 & 32 \\
    \bottomrule
    \end{tabular}}
    \end{center}
    \end{table}

    The best parameters found by Optuna for the GA with serialised evaluations (sGA) and parallelisd evaluations (pGA) on MKP, QAP and TSP instances are presented in Tables \ref{tb:gamkp}, \ref{tb:gaqap} and \ref{tb:gatsp}. In these tables, crossover types $ctype$ 1-p, 2-p, exp, ux and ox are used to denote one-point, two-point, exponential, uniform, and order crossovers respectively. Where duplicates are eliminated; $dup$ is set to T, otherwise; $dup$ is set to F. Parameters presented in Tables \ref{tb:gamkp} - \ref{tb:gatsp} are used for all experiments in later sections.
    
    Although a population size of up to $10n$ was explored, Optuna selected values $p \leq 7n$ for MKP, $p < 10n$ for QAP and $p < 4n$ for TSP instances. Although Optuna consistently selected the same crossover type (\textit{order}) for TSP and QAP instances, the choice was more random on MKP instances. This may be because the problems are smaller, and the GA is therefore less sensitive to the crossover type parameter. Furthermore, eliminating duplicate solutions also seemed to be more favoured when the sGA was applied to TSP and QAP instances but less so in the pGA. For MKP instances, there are no clear indications whether removing duplicates is consistently a good choice. GA variants sGA and pGA seem less sensitive to crossover probability ($a$) and mutation rate ($b$) as this varied widely for all problem categories. For MKP instances, $b$ values were all lower than 0.15 while for QAP and TSP instances, larger values closer to 0.2 were also returned.

 \begin{table}[h!]
    \begin{center}
    \caption{Optuna: selected GA parameters for MKP instances}
    \label{tb:gamkp}
    \resizebox{0.8\columnwidth}{!}{
    \begin{tabular}{@{}ccccccccccc@{}}
    \toprule
    \multirow{3}{*}{Instance} & \multicolumn{10}{c}{Values} \\ \cmidrule(l){2-11} 
     & \multicolumn{5}{c}{sGA} & \multicolumn{5}{c}{pGA} \\ \cmidrule(l){2-11} 
     & $p$ & $c\-type$ & $dup$ & $a$ & $b$ &$p$ & $c\-type$ & $dup$ & $a$ & $b$  \\ \midrule
    weing1 & 67 & exp & T & 0.7129 & 0.0953 & 102 & ux & T & 0.6766 & 0.1746  \\ \midrule
    weing2 & 95 & exp & T & 0.7186 & 0.0283 & 50 & ux & T & 0.8532  & 0.0363 \\ \midrule
    weing3 & 146 & 1-p & F & 0.8333 & 0.1339 & 173  & 2-p & F & 0.5964 & 0.1659 \\ \midrule
    weing4 & 147 & exp & T & 0.6059 & 0.0357 & 176 & 2-p & F & 0.7460  & 0.1946 \\ \midrule
    weing5 & 170 & 2-p & T & 0.8660 & 0.14408 & 158 & ux & F & 0.6558 & 0.1436 \\ \midrule
    weing6 & 53 & 1-p & T & 0.7039 & 0.02567 & 93 & ux & T & 0.5418 & 0.0017 \\ \midrule
    weing7 & 175 & ux & F & 0.7514 & 0.0017 & 111 & 2-p & F & 0.7166 & 0.0147 \\ \midrule
    weing8 & 183 & ux & F & 0.7230 & 0.0835 & 106 & 2-p & F & 0.8467 &  0.0322\\ \bottomrule
    \end{tabular}}
    \end{center}
    \end{table}

 \begin{table}[h!]
    \begin{center}
    \caption{Optuna: selected GA parameters for QAP instances}
    \label{tb:gaqap}
     \resizebox{0.8\columnwidth}{!}{
    \begin{tabular}{@{}ccccccccccc@{}}
    \toprule
    \multirow{3}{*}{Instance} & \multicolumn{10}{c}{Values} \\ \cmidrule(l){2-11} 
     & \multicolumn{5}{c}{sGA} & \multicolumn{5}{c}{pGA} \\ \cmidrule(l){2-11} 
     & $p$ & $c\-type$ & $dup$ & $a$ & $b$ &$p$ & $c\-type$ & $dup$ & $a$ & $b$  \\ \midrule
      had12& 23& ox& T & 0.8348& 0.1592 & 43 & ox & T & 0.7279 & 0.0201\\ \midrule
    had14& 62& ox& F & 0.7304& 0.0572 & 87 & ox & F & 0.8225 & 0.1075\\ \midrule
    had16& 72& ox& F & 0.8935& 0.0461 & 52 & ox & T & 0.5018 & 0.1941\\ \midrule
    had18& 38& ox& F & 0.7868& 0.1382 & 50 & ox & T & 0.8254 & 0.0742\\ \midrule
    had20& 30& ox& F& 0.7015& 0.1234 & 48 & ox & T & 0.5811 & 0.0757\\ \midrule
    rou12& 58& ox& T& 0.6763& 0.0019 & 110 & ox & F & 0.7755 & 0.1304 \\ \midrule
    rou15& 31& ox& F& 0.7025& 0.1108 & 35 & ox & F & 0.8280 & 0.0635 \\ \midrule
    rou20& 48& ox& F& 0.6937& 0.0831 & 29 & ox & F & 0.6279 & 0.1982 \\ \midrule
    tai40a& 40& ox& T& 0.8226& 0.1319 & 41 & ox & T & 0.5901 & 0.0869\\ \midrule
    tai40b& 41& ox& F & 0.7132& 0.1952 & 95 & ox & T & 0.6597 & 0.1303 \\ 
       \bottomrule
    \end{tabular}}
    \end{center}
    \end{table}

 \begin{table}[h!]
    \begin{center}
    \caption{Optuna: selected GA parameters for TSP instances}
    \label{tb:gatsp}
        \resizebox{0.8\columnwidth}{!}{
    \begin{tabular}{@{}ccccccccccc@{}}
    \toprule
    \multirow{3}{*}{Instance} & \multicolumn{10}{c}{Values} \\ \cmidrule(l){2-11} 
     & \multicolumn{5}{c}{sGA} & \multicolumn{5}{c}{pGA} \\ \cmidrule(l){2-11} 
     & $p$ & $c\-type$ & $dup$ & $a$ & $b$ &$p$ & $c\-type$ & $dup$ & $a$ & $b$  \\ \midrule
    bayg29& 30& ox& F& 0.6746& 0.1039 & 30 & ox & F & 0.5528 & 0.1660 \\ \midrule
    bays29& 30& ox& F& 0.5641& 0.1988 & 30 & ox & T & 0.5024 & 0.1205 \\ \midrule 
    berlin52& 52& ox& F& 0.8112& 0.1416 & 54 & ox & T & 0.5000 & 0.1520 \\ \midrule
    brazil58& 65& ox& F& 0.6834& 0.1129 & 57 & ox & F & 0.7465 & 0.0068\\ \midrule
    dantzig42& 43& ox& F& 0.8986& 0.0002 & 41 & ox & F & 0.7464 & 0.1607  \\ \midrule
    fri26& 25& ox& F& 0.8409& 0.0564 & 26 & ox & F & 0.6494 & 0.1196\\ \midrule
    gr17& 64& ox& F& 0.5015& 0.0284 & 64 & ox & F & 0.8537 & 0.1182 \\ \midrule
    gr21& 32& ox& F& 0.5014& 0.0439 & 22 & ox & F & 0.6891 & 0.0642 \\ \midrule
    gr24& 39& ox& F& 0.5917& 0.1853 & 34 & ox & F & 0.5088 & 0.1127\\ \midrule
    st70& 70& ox& F& 0.7294& 0.0568 & 69 & ox & F & 0.6420 & 0.0997\\ 
   \bottomrule
    \end{tabular}}
    \end{center}
    \end{table}

    \subsubsection{Digital Annealer}
    Although the DA has its own inbuilt automatic parameter tuning feature, we used Optuna to tune some of the more important parameters so we can make a fairer comparison with the GA. The parameters tuned are:

    \textbf{gs\_level}: setting a lower value reduces the number of iterations required before restarting the search and therefore results in higher diversification of solutions. Possible range of values are 0 to 100, default value is 5. 
    
    \textbf{gs\_cutoff}: reducing this value results in an earlier restart of the search if no improvement in objective function is seen, resulting in higher diversification of solutions. Possible range of values are 0 to $1,000,000$ while default value is 8000.

    \textbf{num\_run}: this parameter defines the number of parallel attempts of each group. Increasing num\_run results in higher diversification of solutions. Possible range of values are 1 to 16, default value is 16.

    \textbf{num\_group}: this parameter defines the number of groups of parallel attempts. A higher num\_group results in reducing the sensitivity to randomness. Possible range of values are 1 to 16, default value is 1.  Note that num\_run $\times$ num\_group specifies the number of parallel attempts.

    \textbf{penalty weight} ($\alpha$): the DA solves QUBO matrix $Q = C + \alpha \cdot G$. $Q$ is a weighted aggregate of QUBOs representing objective function ($C$) and constraint function ($G$). To derive $\alpha$, we did not use the automatic tuning of the DA, we used method presented in \cite{verma2020penalty} defined in Eq. \eqref{eq:wvermaa}. This is so that we can assess the algorithm on time spent finding a solution and not time spent finding parameters.
         
         \begin{equation}
        \label{eq:wvermaa}
        \alpha = max\Biggl\{ -C_{i,i} - \sum_{\substack{j=1\\ j \neq i}}^{n}  \min\{C_{i,j}, 0\},\  
         C_{i,i}
        + \sum_{\substack{j=1\\ j \neq i}}^{n} \max\{C_{i,j}, 0\}\  \forall\ i \in  \left [ 1,n \right ] \Biggr\} 
        \end{equation}

    Optuna was used to explore the full range of parameters values for gs\_level, gs\_cutoff, num\_run and num\_group. We allowed a maximum of 1 second, 20 repeated runs and 30 trials as done for the GA. The DA was however set to terminate earlier if it finds the optimal before 1 seconds. Also, to save computation time, Optuna was not executed on instances where the DA found optimal within 1 second using its default settings. Parameters used by the DA in the experiments comparing the DA with GA are presented in Tables \ref{tb:damkp} (MKP), \ref{tb:daqap} (QAP) and \ref{tb:datsp} (TSP).

     \begin{table}[h!]
    \begin{center}
    \caption{Selected DA parameters for MKP instances}
    \label{tb:damkp}
    \resizebox{0.5\columnwidth}{!}{
    \begin{tabular}{@{}ccccc@{}}
    \toprule
    \multirow{2}{*}{\textbf{Instance}} &  \multicolumn{4}{c}{\textbf{Values}} \\ \cmidrule(l){2-5} 
     &  gs\_level & gs\_cutoff &  num\_run & num\_group \\ \midrule
    weing1 & 12 & 277,512 & 14 & 2 \\ \midrule
    weing2 & default & default & default & default\\ \midrule
    weing3 & 35 & 984,915 &  5 & 1\\ \midrule
    weing4& default & default & default & default\\ \midrule
    weing5 & default & default & default & default\\ \midrule
    weing6 & 43 & 791,905 & 5 & 11 \\ \midrule
    weing7 & default & default & default & default\\ \midrule
    weing8 &100 & 312191 & 8 & 14\\ 
   \bottomrule
    \end{tabular}}
    \end{center}
    \end{table}

     \begin{table}[h!]
    \begin{center}
    \caption{Selected DA parameters for QAP instances}
    \label{tb:daqap}
      \resizebox{0.5\columnwidth}{!}{
    \begin{tabular}{@{}ccccc@{}}
    \toprule
    \multirow{2}{*}{\textbf{Instance}} &  \multicolumn{4}{c}{\textbf{Values}} \\ \cmidrule(l){2-5} 
     &  gs\_level & gs\_cutoff &  num\_run & num\_group \\ \midrule
    had12 & default & default & default& default\\ \midrule
    had14 & default & default & default& default\\ \midrule
    had16 & default & default & default& default\\ \midrule
    had18 & default & default & default& default\\ \midrule
    had20 & default & default & default& default\\ \midrule
    rou12 & default & default & default& default\\ \midrule
    rou15 & default & default & default& default\\ \midrule
    rou20 & default & default & default& default\\ \midrule
    tai40a & 67 & 621,145 & 7 & 14 \\ \midrule
    tai40b & 48 & 383,240 & 4 & 10 \\ 
   \bottomrule
    \end{tabular}}
    \end{center}
    \end{table}

   \begin{table}[h!]
    \begin{center}
    \caption{Selected DA parameters for TSP instances}
    \label{tb:datsp}
      \resizebox{0.5\columnwidth}{!}{
    \begin{tabular}{@{}ccccc@{}}
    \toprule
    \multirow{2}{*}{\textbf{Instance}} &  \multicolumn{4}{c}{\textbf{Values}} \\ \cmidrule(l){2-5} 
     &  gs\_level & gs\_cutoff &  num\_run & num\_group \\ \midrule
    bayg29 & default & default & default & default\\ \midrule
    bays29 & default & default & default & default\\ \midrule
    berlin52  & 91& 741,373& 10& 16\\ \midrule
    brazil58  & 24& 1,243& 5& 13\\ \midrule
    dantzig42  & 41& 249,453& 15& 15\\ \midrule
    fri26 & default & default & default  & default\\ \midrule
    gr17 & default & default & default  & default\\ \midrule
    gr21 & default & default & default  & default\\ \midrule
    gr24 & default & default & default  & default\\ \midrule
    st70 & 20 & 345,700 & 5& 4 \\
   \bottomrule
    \end{tabular}}
    \end{center}
    \end{table}
\section{Results}
\label{sec:res}
In this section, we compare the DA with GA on MKP, QAP and TSP instances with stopping criteria 1, 2, 5 and 10 second(s) run times. Parameters presented in Section \ref{sec:ex} were used when executing the DA and GA. We note that both algorithms found feasible solutions in 1 second, all results reported in this section are therefore for feasible solutions only. 
We measure statistical significance using the student t-test.

\subsection{Comparing sGA and pGA}
In Tables \ref{tb:mkpgaga}, \ref{tb:qapgaga} and \ref{tb:tspgaga}, we compare GA which uses serialised evaluations (sGA) with one that uses parallelised evaluations (pGA). We show that GA with serialised evaluations (sGA) produced similar or better average fitness compared to the GA which uses parallelised evaluations. Although preliminary results suggest that using JIT led to significant speed up in both implementations, it was however the case that sGA was able to exploit speedup provided by the numba library \cite{lam2015numba} better than pGA. We therefore use sGA for further experiments. The results for GA presented in further sections of this study therefore refers to sGA.  

\begin{table}[tbh]
\centering
\caption{Comparing serialised and parallelised GA on MKP: Average fitness across 20 runs within 1 second run time. Smaller values are better. Optimal average fitness values are displayed in bold. Asterisk (*) is appended where one algorithm is significantly better than the other}\label{tb:mkpgaga}
 \resizebox{0.5\columnwidth}{!}{
\begin{tabular}{@{}cccc@{}}
\toprule
\multirow{2}{*}{Instances} & \multirow{2}{*}{Optimal} & \multicolumn{2}{c}{\begin{tabular}[c]{@{}c@{}}Average Fitness \\ (Standard deviation)\end{tabular}} \\ \cmidrule(l){3-4} 
 &  & sGA & pGA \\ \midrule
weing1 & 141,278 &  141,272 (28) & 141,277 (4)  \\ \midrule
weing2 & 130,883 & \textbf{130,883} (0) & \textbf{130,883} (0)\\ \midrule
weing3 & 95,677 & \textbf{95,677} (0) & 95,675 (11)   \\ \midrule
weing4 & 119,337 & \textbf{119,337} (0) & \textbf{119,337} (0)  \\ \midrule
weing5 & 98,796& \textbf{98,796} (0) & \textbf{98,796} (0)  \\ \midrule
weing6 & 130,623 & \textbf{130,623} (0) & 130,597 (113) \\ \midrule
weing7 & 1,095,445 & 1,095,382 (0)* & 1,091,203 (2471)  \\ \midrule
weing8 & 624,319  & 621,086 (0)* & 570,440 (10,204)  \\ 
\bottomrule
\end{tabular}}
\end{table}

\begin{table}[tbh]
\centering
\caption{Comparing serialised and parallelised GA on QAP: Average fitness across 20 runs within 1 second run time. Smaller values are better. Optimal average fitness values are displayed in bold. Asterisk (*) is appended where one algorithm is significantly better than the other}\label{tb:qapgaga}
 \resizebox{0.65\columnwidth}{!}{
\begin{tabular}{@{}cccc@{}}
\toprule
\multirow{2}{*}{Instances} & \multirow{2}{*}{Optimal} & \multicolumn{2}{c}{\begin{tabular}[c]{@{}c@{}}Average Fitness \\ (Standard deviation)\end{tabular}} \\ \cmidrule(l){3-4} 
 &  & sGA & pGA \\ \midrule
had12 & 1,652 &  \textbf{1,652} (0) & \textbf{1,652} (0)\\ \midrule
had14 & 2,724 &  \textbf{2,724} (0) & \textbf{2,724} (0)  \\ \midrule
had16 & 3,720 &  \textbf{3,720} (0)* & 3,723 (3)\\ \midrule
had18 & 5,358 & 5,366 (0)* & 5,375 (4) \\ \midrule
had20 & 6,922 &  6,940 (0)* & 6,969 (6)  \\ \midrule
rou12 & 235,528  & 238,152 (80)* &  236,340 (3,540)\\ \midrule
rou15 & 354,210  & 368,444 (0)* &  363,538 (2,134) \\ \midrule
rou20 & 725,522  & 763,280 (0)* & 764,064 (4,262) \\ \midrule
tai40a  & 3,139,370  & 3,482,323 (608)* & 3,494,273 (265) \\ \midrule
tai40b & 637,250,948 & 672,808,717 (102,353)* & 716,630,961 (7,303,076)\\ 
\bottomrule
\end{tabular}}
\end{table}

\begin{table}[tbh]
\centering
\caption{Comparing serialised and parallelised GA on TSP: Average fitness across 20 runs within 1 second run time. Smaller values are better. Optimal average fitness values are displayed in bold. Asterisk (*) is appended where one algorithm is significantly better than the other}\label{tb:tspgaga}
 \resizebox{0.4\columnwidth}{!}{
\begin{tabular}{@{}cccc@{}}
\toprule
\multirow{2}{*}{Instances} & \multirow{2}{*}{Optimal} & \multicolumn{2}{c}{\begin{tabular}[c]{@{}c@{}}Average Fitness \\ (Standard deviation)\end{tabular}} \\ \cmidrule(l){3-4} 
 &  & sGA & pGA \\ \midrule
bayg29 & 1,610 & 1,673 (8)* & 1,791 (6) \\ \midrule
bays29 & 2,020 & 2,050 (24)* & 2,142 (55) \\ \midrule
berlin52 & 7,542 &  11,228 (66)* & 11,741 (253) \\ \midrule
brazil58 & 25,395 &  40,882 (333)*  & 50,730 (828) \\ \midrule
dantzig42 & 699 &  908 (6)* & 1,099 (6) \\\midrule
fri26 & 937 & 976 (0)*  & 992 (17) \\ \midrule
gr17 & 2,085 & 2,090 (0)*  & 2,129 (2)  \\ \midrule
gr21 & 2,707 &  2,801 (0)*  & 2,803 (0)  \\ \midrule
gr24 & 1,272  & 1,283 (0)*  & 1,401 (12)  \\ \midrule
st70 & 675  &  1,541 (0)* & 1,803 (23)  \\ 
\bottomrule
\bottomrule
\end{tabular}}
\end{table}

\subsection{Comparing GA and DA}
\subsubsection{Multi-dimensional Knapsack Problem}

\begin{table}[h!]
\begin{center}
\caption{Comparing GA and DA on MKP: Average fitness across 20 runs within 1, 2, 5 and 10 second(s) run times. Larger values are better. Optimal average fitness values are displayed in bold. Asterisk (*) is appended where one algorithm is significantly better than the other}\label{tb:mkpdaga}
\resizebox{ 0.7\columnwidth}{!}{
\begin{tabular}{@{}cccccc@{}}
\toprule
\multirow{2}{*}{\textbf{Instance}} & \multirow{2}{*}{\textbf{Optimal}} & \multicolumn{4}{c}{\textbf{Average Fitness (Standard deviation Fitness)}} \\ \cmidrule(l){3-6} 
 &  & \textbf{DA (1s)} & \textbf{GA (1s)} & \textbf{DA (2s) } & \textbf{GA (2s)} \\ \midrule
weing1 & 141,278 & \textbf{141,278} (0) &  141,272 (28) & \textbf{141,278} (0)  & \textbf{141,278} (0) \\ \midrule
weing2 & 130,883 & \textbf{130,883} (0) & \textbf{130,883} (0) & \textbf{130,883} (0) & \textbf{130,883} (0)\\ \midrule
weing3 & 95,677 & \textbf{95,677} (0) & \textbf{95,677} (0) & \textbf{95,677} (0) & \textbf{95,677} (0) \\ \midrule
weing4 & 119,337 & \textbf{119,337} (0) & \textbf{119,337} (0) & \textbf{119,337} (0) & \textbf{119,337} (0) \\ \midrule
weing5 & 98,796& \textbf{98,796} (0) & \textbf{98,796} (0) & \textbf{98,796} (0) & \textbf{98,796} (0)\\ \midrule
weing6 & 130,623 & \textbf{130,623} (0) & \textbf{130,623} (0)  & \textbf{130,623} (0)   & \textbf{130,623} (0) \\ \midrule
weing7 & 1,095,445 & \textbf{1,095,445} (0)* & 1,095,382 (0) & \textbf{1,095,445} (0)* & 1,095,382 (0)  \\ \midrule
weing8 & 624,319  & 621,086 (0) & 621,086 (0) & 621,086 (0)  & 621,086 (0) \\ 
\bottomrule
\end{tabular}}
\resizebox{ 0.55\columnwidth}{!}{
\begin{tabular}{@{}cccc@{}}
\toprule
\multirow{2}{*}{\textbf{Instance}} & \multirow{2}{*}{\textbf{Optimal}} & \multicolumn{2}{c}{\textbf{Average Fitness (Standard deviation Fitness)}} \\ \cmidrule(l){3-4} 
 &  & \textbf{DA (5s or 10s)} & \textbf{GA (5s or 10s)} \\ \midrule
weing1 & 141,278 & \textbf{141,278} (0) &  \textbf{141,278} (0)  \\ \midrule
weing2 & 130,883 & \textbf{130,883} (0) & \textbf{130,883} (0) \\ \midrule
weing3 & 95,677 & \textbf{95,677} (0) & \textbf{95,677} (0) \\ \midrule
weing4 & 119,337 & \textbf{119,337} (0) & \textbf{119,337} (0)  \\ \midrule
weing5 & 98,796& \textbf{98,796} (0) & \textbf{98,796} (0) \\ \midrule
weing6 & 130,623 & \textbf{130,623} (0) & \textbf{130,623} (0)   \\ \midrule
weing7 & 1,095,445 & \textbf{1,095,445} (0)* & 1,095,382 (0) \\ \midrule
weing8 & 624,319  & \textbf{624,319} (0)*  & 621,086 (0)  \\ 
\bottomrule
\end{tabular}}
\end{center}
\end{table}

The DA was compared to the GA on MKP instances. Results based on 1, 2, 5 and 10 seconds time limits are presented in Table \ref{tb:mkpdaga}. The results shown for the DA is based on the inequality representation of the constraint. We note that the GA did not improve on any of the instances between 5 and 10 seconds time limit. The DA reached optimal on 9 of 10 instances in 1 second and all MKP instances within 5 seconds. GA reached optimal on 5 instances in 1 second 6 instances in 5 seconds. The average fitness achieved by DA was significantly better than that of the GA on 2 instance within the maximum time limit of 10 seconds. Based on these results, we can conclude that overall better performance was achieved by DA compared to the GA.

\subsubsection{Quadratic Assignment Problem}
Performance of the DA and GA are compared on QAP instances within 1, 2, 5 and 10 second(s) in Table \ref{tb:qapdaga}. We show that the DA reached optimal on all 10 instances within 1 second. Results of running the DA for 2, 5 and 10 seconds are therefore not shown in Table \ref{tb:qapdaga}. The GA reached optimal on 3 out of 10 instances within 10 seconds. With the exception of the 3 instances where both DA and GA reached optimal, DA achieved significantly better average fitness in 1 second than GA executed for 1, 2, 5 or 10 seconds. We can therefore suggest that a speed up of more than ten times could be achieved using the DA compared to using the GA on QAP instances.

\begin{table}[htb]
\begin{center}
\caption{Comparing GA and DA on QAP: Average fitness across 20 runs within 1, 2, 5 and 10 second(s) run times. Smaller values are better. Optimal average fitness values are displayed in bold. Asterisk (*) is appended where one algorithm is significantly better than the other}\label{tb:qapdaga}
\resizebox{ 0.6\columnwidth}{!}{
\begin{tabular}{@{}cccc@{}}
\toprule
\multirow{2}{*}{\textbf{Instance}} & \multirow{2}{*}{\textbf{Optimal}} & \multicolumn{2}{c}{\textbf{Average Fitness (Standard deviation Fitness)}} \\ \cmidrule(l){3-4} 
 &  & \textbf{DA (1s)} & \textbf{GA (1s)} \\ \midrule
had12 & 1,652 & \textbf{1,652} (0) & \textbf{1,652} (0) \\ \midrule
had14 & 2,724 & \textbf{2,724} (0) & \textbf{2,724} (0)  \\ \midrule
had16 & 3,720 & \textbf{3,720} (0) & \textbf{3,720} (0) \\ \midrule
had18 & 5,358 & \textbf{5,358} (0)* & 5,366 (0)  \\ \midrule
had20 & 6,922 & \textbf{6,922} (0)* & 6,940 (0)  \\ \midrule
rou12 & 235,528 & \textbf{235,528} (0)* & 238,152 (80) \\ \midrule
rou15 & 354,210 & \textbf{354,210} (0)* & 368,444 (0) \\ \midrule
rou20 & 725,522 & \textbf{725,522} (0)* & 763,280 (0) \\ \midrule
tai40a  & 3,139,370 & \textbf{3,139,370} (0)* & 3,482,323 (608) \\ \midrule
tai40b & 637,250,948 & \textbf{637,250,948} (0)* & 672,808,717 (102,353) \\ 
\bottomrule
\end{tabular}}
\resizebox{ 0.6\columnwidth}{!}{
\begin{tabular}{@{}cccc@{}}
\toprule
\multirow{2}{*}{\textbf{Instance}} &  \multicolumn{3}{c}{\textbf{Average Fitness (Standard deviation Fitness)}} \\ \cmidrule(l){2-4} 
 & \textbf{GA (2s)} & \textbf{GA (5s)} & \textbf{GA (10s)} \\ \midrule
had12 & \textbf{1,652} (0) & \textbf{1,652} (0) & \textbf{1,652} (0)  \\ \midrule
had14 & \textbf{2,724} (0) & \textbf{2,724} (0) & \textbf{2,724} (0) \\ \midrule
had16 & \textbf{3,720} (0) & \textbf{3,720} (0) & \textbf{3,720} (0) \\ \midrule
had18 &  5,364 (0) & 5,364 (0) & 5,364 (0) \\ \midrule
had20 & 6,936 (0) & 6,936 (0) & 6,936 (0) \\ \midrule
rou12 & 238,134 (0)  & 238,134 (0) & 238,134 (0) \\ \midrule
rou15 & 368,444 (0) (0) & 368,444 (0) & 368,444 (0) \\ \midrule
rou20 & 760,200 (0) & 749,994 (0) & 749,994 (0) \\ \midrule
tai40a  & 3,425,484 (2,275) & 3,395,666 (181) & 3,374,710 (0) \\ \midrule
tai40b & 668,035,169 (50,021) & 665,485,935 (0) & 662,447,324 (0) \\ 
\bottomrule
\end{tabular}}
\end{center}
\end{table}

\subsubsection{Travelling Salesman Problem}
Performance of DA and GA are compared on TSP instances in Table \ref{tb:tspdaga}. We show that the DA reached optimal on 6 out of 10 instances while the GA reached optimal on 1 out of 10 instances. The DA consistently produced similar or better results than the GA when executed for 1 second and 2 seconds. Although the DA achieved significantly better average fitness than GA on instances $berlin52$, $brazil58$ and $st70$ when both algorithms were executed for 2 seconds, GA improved its performance quicker on these instances and was able to show better performance within 10 seconds. At the end of 10 seconds, DA achieved better average fitness than GA on 6 TSP instances, the same average fitness on 1 TSP instances while the GA achieved better average fitness than DA on 2 TSP instances.

\begin{table}[tbh]
\begin{center}
\caption{Comparing GA and DA on TSP: Average fitness across 20 runs within 1, 2, 5 and 10 second(s) run times. Smaller values are better. Optimal average fitness values are displayed in bold. Asterisk (*) is appended where one algorithm is significantly better than the other}\label{tb:tspdaga}
\resizebox{0.6 \columnwidth}{!}{
\begin{tabular}{@{}cccccc@{}}
\toprule
\multirow{2}{*}{\textbf{Instance}} & \multirow{2}{*}{\textbf{Optimal}} & \multicolumn{4}{c}{\textbf{Average Fitness (Standard deviation Fitness)}} \\ \cmidrule(l){3-6} 
 &  & \textbf{DA (1s)} & \textbf{GA (1s)} & \textbf{DA (2s) } & \textbf{GA (2s)} \\ \midrule
bayg29 & 1,610 & \textbf{1,610} (0)* & 1,673 (8) &  \textbf{1,610} (0)* &  1,671 (0) \\ \midrule
bays29 & 2,020 & \textbf{2,020} (0)* & 2,050 (24)   &  \textbf{2,020} (0)*   &  2,028 (0) \\ \midrule
berlin52 & 7,542 & 8,159 (0)* &  11,228 (66) &  8,159 (0)* &  9,990 (4) \\ \midrule
brazil58 & 25,395 & 28,921 (479)* & 40,882 (333)  & 28,480 (93)*  &  33,519 (8) \\ \midrule
dantzig42 & 699 &  710 (0)* &  908 (6) & 710 (0)* & 820 (0) \\\midrule
fri26 & 937 & \textbf{937} (0)* & 976 (0)  & \textbf{937} (0)* & 957 (0)  \\ \midrule
gr17 & 2,085 & \textbf{2,085} (0)* & 2,090 (0)  & \textbf{2,085} (0)* & 2,090 (0) \\ \midrule
gr21 & 2,707 & \textbf{2,707} (0)* & 2,801 (0)  & \textbf{2,707} (0)* & 2,801 (0) \\ \midrule
gr24 & 1,272 & \textbf{1,272} (0)* & 1,283 (0)  & \textbf{1,272} (0) &  \textbf{1,272} (0) \\ \midrule
st70 & 675 &  793 (0)*  &  1,541 (0)&  793 (0)*  &  1,211 (13) \\ 
\bottomrule
\end{tabular}}
\resizebox{0.6 \columnwidth}{!}{
\begin{tabular}{@{}cccccc@{}}
\toprule
\multirow{2}{*}{\textbf{Instance}} & \multirow{2}{*}{\textbf{Optimal}} & \multicolumn{4}{c}{\textbf{Average Fitness (Standard deviation Fitness)}} \\ \cmidrule(l){3-6} 
 &  & \textbf{DA (5s)} & \textbf{GA (5s)} & \textbf{DA (10s) } & \textbf{GA (10s)} \\ \midrule
bayg29 & 1,610 & \textbf{1,610} (0)* & 1,635 (0)  &  \textbf{1,610} (0)* & 1,628 (0)  \\ \midrule
bays29 & 2,020 & \textbf{2,020} (0)* & 2,028 (0)  &  \textbf{2,020} (0)*   &  2,028 (0) \\ \midrule
berlin52 & 7,542 & 8,159 (0)  &  8,026 (0)*  & 8,159 (0)  & 7,567 (0)*  \\ \midrule
brazil58 & 25,395 &  26,981 (5)  & 26,972 (0)*  & 26,508 (201) & 25,723 (46)*  \\ \midrule
dantzig42 & 699 &  710 (0)* & 718 (0)   & 700 (0)*  & 718 (0) \\\midrule
fri26 & 937 & \textbf{937} (0)* &  953 (0)   & \textbf{937} (0)* & 953 (0)  \\ \midrule
gr17 & 2,085 & \textbf{2,085} (0)* &  2,090 (0) & \textbf{2,085} (0)* &  2,090 (0)\\ \midrule
gr21 & 2,707 & \textbf{2,707} (0)* &  2,801 (0) & \textbf{2,707} (0)* & 2,801 (0) \\ \midrule
gr24 & 1,272 & \textbf{1,272} (0) &   \textbf{1,272} (0) & \textbf{1,272} (0) &  \textbf{1,272} (0) \\ \midrule
st70 & 675 &  760 (7)* & 832 (1) & 754 (0)  & 748 (0)*  \\ 

\bottomrule
\end{tabular}}
\end{center}
\end{table}

Both the DA and GA struggled to improve their performance after 1 second on many instances. This might be attributed to the fact that Optuna was set to find the best parameters for 1 second time limit. Optuna may therefore have chosen parameters that led to slower convergence but eventual better performance if allowed to run for longer.

At 1 second time limit, compared to the GA, the DA achieved similar or better average fitness on all instances. This is also true for MKP and QAP instances when the algorithms were allowed to execute for 10 seconds. However, although the DA was better than the GA on 6 TSP instances, there were 3 instances where the GA was able to reach better solutions than DA when allowed to run for 10 seconds. Overall, the highest performance improvement in favour of DA was on QAP instances. This may be because the QAP is naturally quadratic.

\section{Conclusion}
\label{sec:con}
In this work, we compared the performance of an evolutionary algorithm, GA with a quantum-inspired hardware solver, DA. This research set out to assess whether the DA could perform better than GA implemented on high performance computer or vice versa. We show that even with fast implementation of the GA, the DA often achieved better results. We see a speedup of more than 10 on QAP instances. The different in performance was however not as high for MKP instances. On TSP instances, DA was often better but there were instances where GA improved quicker than DA to achieve better results.

This work presents many future directions:
\begin{itemize}
    \item a framework for fairer comparison between algorithms that use different hardware and different approach of fitness evaluations
    
    \item compare a wider range of quantum or quantum-inspired algorithms with a wider range of evolutionary algorithms
    
    \item more extensive hyper-parameter tuning ensuring each algorithm is evaluated on parameters that are well suited for the problem solved and the time limit set
\end{itemize}

% \begin{acks}
% M.\@ L\'opez-Ib\'a\~nez is a ``Beatriz Galindo'' Senior Distinguished Researcher (BEAGAL 18/00053) funded by the Spanish Ministry of Science and Innovation (MICINN).
% \end{acks}

\bibliographystyle{unsrtnat}
\bibliography{References}

\begin{thebibliography}{33}
\providecommand{\natexlab}[1]{#1}
\providecommand{\url}[1]{\texttt{#1}}
\expandafter\ifx\csname urlstyle\endcsname\relax
  \providecommand{\doi}[1]{doi: #1}\else
  \providecommand{\doi}{doi: \begingroup \urlstyle{rm}\Url}\fi

\bibitem[McGeoch and Farr{\'e}(2020)]{mcgeoch2020d}
Catherine McGeoch and Pau Farr{\'e}.
\newblock The d-wave advantage system: An overview.
\newblock \emph{D-Wave Systems Inc., Burnaby, BC, Canada, Tech. Rep}, 2020.

\bibitem[Chow et~al.(2021)Chow, Dial, and Gambetta]{chow2021ibm}
Jerry Chow, Oliver Dial, and Jay Gambetta.
\newblock Ibm quantum breaks the 100-qubit processor barrier.
\newblock \emph{IBM Research Blog}, 2021.
\newblock URL
  \url{https://research.ibm.com/blog/127-qubit-quantum-processor-eagle}.

\bibitem[Hiroshi et~al.(2021)Hiroshi, Junpei, Noboru, and Toshiyuki]{da3}
Nakayama Hiroshi, Koyama Junpei, Yoneoka Noboru, and Miyazawa Toshiyuki.
\newblock Third generation digital annealer technology, 2021.
\newblock URL
  \url{https://www.fujitsu.com/jp/documents/digitalannealer/researcharticles/DA_WP_EN_20210922.pdf}.

\bibitem[Aramon et~al.(2019)Aramon, Rosenberg, Valiante, Miyazawa, Tamura, and
  Katzgraber]{aramon2019physics}
Maliheh Aramon, Gili Rosenberg, Elisabetta Valiante, Toshiyuki Miyazawa,
  Hirotaka Tamura, and Helmut~G Katzgraber.
\newblock Physics-inspired optimization for quadratic unconstrained problems
  using a digital annealer.
\newblock \emph{Frontiers in Physics}, 7:\penalty0 48, 2019.

\bibitem[Katoch et~al.(2021)Katoch, Chauhan, and Kumar]{katoch2021review}
Sourabh Katoch, Sumit~Singh Chauhan, and Vijay Kumar.
\newblock A review on genetic algorithm: past, present, and future.
\newblock \emph{Multimedia Tools and Applications}, 80\penalty0 (5):\penalty0
  8091--8126, 2021.

\bibitem[Storn and Price(1997)]{storn1997differential}
Rainer Storn and Kenneth Price.
\newblock Differential evolution--a simple and efficient heuristic for global
  optimization over continuous spaces.
\newblock \emph{Journal of global optimization}, 11\penalty0 (4):\penalty0
  341--359, 1997.

\bibitem[Larra{\~n}aga and Lozano(2001)]{larranaga2001estimation}
Pedro Larra{\~n}aga and Jose~A Lozano.
\newblock \emph{Estimation of distribution algorithms: A new tool for
  evolutionary computation}, volume~2.
\newblock Springer Science \& Business Media, 2001.

\bibitem[Salkin and De~Kluyver(1975)]{salkin1975knapsack}
Harvey~M Salkin and Cornelis~A De~Kluyver.
\newblock The knapsack problem: a survey.
\newblock \emph{Naval Research Logistics Quarterly}, 22\penalty0 (1):\penalty0
  127--144, 1975.

\bibitem[Chu and Beasley(1998)]{chu1998genetic}
Paul~C Chu and John~E Beasley.
\newblock A genetic algorithm for the multidimensional knapsack problem.
\newblock \emph{Journal of heuristics}, 4\penalty0 (1):\penalty0 63--86, 1998.

\bibitem[Wang et~al.(2012)Wang, Wang, and Xu]{wang2012effective}
Ling Wang, Sheng-yao Wang, and Ye~Xu.
\newblock An effective hybrid eda-based algorithm for solving multidimensional
  knapsack problem.
\newblock \emph{Expert Systems with Applications}, 39\penalty0 (5):\penalty0
  5593--5599, 2012.

\bibitem[Diez~García et~al.(2022)Diez~García, Ayodele, and
  Moraglio]{Diez2022Exact}
Marcos Diez~García, Mayowa Ayodele, and Alberto Moraglio.
\newblock Exact and sequential penalty weights in quadratic unconstrained
  binary optimisation with a digital annealer.
\newblock In \emph{Proceedings of the Genetic and Evolutionary Computation
  Conference Companion}, GECCO '22, New York, NY, USA, 2022. Association for
  Computing Machinery.
\newblock ISBN 978-1-4503-9268-6/22/07.
\newblock \doi{10.1145/3520304.3528925}.

\bibitem[Ayodele et~al.(2016)Ayodele, McCall, and
  Regnier-Coudert]{ayodele2016rk}
Mayowa Ayodele, John McCall, and Olivier Regnier-Coudert.
\newblock Rk-eda: A novel random key based estimation of distribution
  algorithm.
\newblock In \emph{International Conference on Parallel Problem Solving from
  Nature}, pages 849--858. Springer, 2016.

\bibitem[Grefenstette et~al.(1985)Grefenstette, Gopal, Rosmaita, and
  Van~Gucht]{grefenstette1985genetic}
John Grefenstette, Rajeev Gopal, Brian Rosmaita, and Dirk Van~Gucht.
\newblock Genetic algorithms for the traveling salesman problem.
\newblock In \emph{Proceedings of the first International Conference on Genetic
  Algorithms and their Applications}, volume 160, pages 160--168. Lawrence
  Erlbaum, 1985.

\bibitem[Tate and Smith(1995)]{tate1995genetic}
David~M Tate and Alice~E Smith.
\newblock A genetic approach to the quadratic assignment problem.
\newblock \emph{Computers \& Operations Research}, 22\penalty0 (1):\penalty0
  73--83, 1995.

\bibitem[Ahuja et~al.(2000)Ahuja, Orlin, and Tiwari]{ahuja2000greedy}
Ravindra~K Ahuja, James~B Orlin, and Ashish Tiwari.
\newblock A greedy genetic algorithm for the quadratic assignment problem.
\newblock \emph{Computers \& Operations Research}, 27\penalty0 (10):\penalty0
  917--934, 2000.

\bibitem[Mi et~al.(2010)Mi, Huifeng, Ming, and Yu]{mi2010improved}
Mei Mi, Xue Huifeng, Zhong Ming, and Gu~Yu.
\newblock An improved differential evolution algorithm for tsp problem.
\newblock In \emph{2010 International Conference on Intelligent Computation
  Technology and Automation}, volume~1, pages 544--547. IEEE, 2010.

\bibitem[Hameed et~al.(2018)Hameed, Aboobaider, Choon, and
  Mutar]{hameed2018improved}
Asaad~Shakir Hameed, Burhanuddin~Mohd Aboobaider, Ngo~Hea Choon, and
  Modhi~Lafta Mutar.
\newblock Improved discrete differential evolution algorithm in solving
  quadratic assignment problem for best solutions.
\newblock \emph{International Journal of Advanced Computer Science and
  Applications}, 9\penalty0 (12):\penalty0 434--439, 2018.

\bibitem[Matsubara et~al.(2020)Matsubara, Takatsu, Miyazawa, Shibasaki,
  Watanabe, Takemoto, and Tamura]{matsubara2020digital}
Satoshi Matsubara, Motomu Takatsu, Toshiyuki Miyazawa, Takayuki Shibasaki,
  Yasuhiro Watanabe, Kazuya Takemoto, and Hirotaka Tamura.
\newblock Digital annealer for high-speed solving of combinatorial optimization
  problems and its applications.
\newblock In \emph{2020 25th Asia and South Pacific Design Automation
  Conference (ASP-DAC)}, pages 667--672. IEEE, 2020.

\bibitem[Ayodele(2022)]{ayodele2022penalty}
Mayowa Ayodele.
\newblock Penalty weights in qubo formulations: Permutation problems.
\newblock In \emph{European Conference on Evolutionary Computation in
  Combinatorial Optimization (Part of EvoStar)}, pages 159--174. Springer,
  2022.

\bibitem[Warren(2020)]{warren2020solving}
Richard~H Warren.
\newblock Solving the traveling salesman problem on a quantum annealer.
\newblock \emph{SN Applied Sciences}, 2\penalty0 (1):\penalty0 1--5, 2020.

\bibitem[Kuramata et~al.(2021)Kuramata, Katsuki, and
  Nakata]{kuramata2021larger}
Michiya Kuramata, Ryota Katsuki, and Kazuhide Nakata.
\newblock Larger sparse quadratic assignment problem optimization using quantum
  annealing and a bit-flip heuristic algorithm.
\newblock In \emph{2021 IEEE 8th International Conference on Industrial
  Engineering and Applications (ICIEA)}, pages 556--565. IEEE, 2021.

\bibitem[Regnier-Coudert and McCall(2014)]{regnier2014factoradic}
Olivier Regnier-Coudert and John McCall.
\newblock Factoradic representation for permutation optimisation.
\newblock In \emph{International Conference on Parallel Problem Solving from
  Nature}, pages 332--341. Springer, 2014.

\bibitem[Baluja(1994)]{baluja1994population}
Shumeet Baluja.
\newblock Population-based incremental learning. a method for integrating
  genetic search based function optimization and competitive learning.
\newblock Technical report, Carnegie-Mellon Univ Pittsburgh Pa Dept Of Computer
  Science, 1994.

\bibitem[Larranaga et~al.(1999)Larranaga, Kuijpers, Murga, Inza, and
  Dizdarevic]{larranaga1999genetic}
Pedro Larranaga, Cindy M.~H. Kuijpers, Roberto~H. Murga, Inaki Inza, and Sejla
  Dizdarevic.
\newblock Genetic algorithms for the travelling salesman problem: A review of
  representations and operators.
\newblock \emph{Artificial intelligence review}, 13\penalty0 (2):\penalty0
  129--170, 1999.

\bibitem[Lucas(2014)]{lucas2014ising}
Andrew Lucas.
\newblock Ising formulations of many np problems.
\newblock \emph{Frontiers in physics}, 2:\penalty0 5, 2014.

\bibitem[Khumalo et~al.(2021)Khumalo, Chieza, Prag, and
  Woolway]{khumalo2021investigation}
Maxine~T Khumalo, Hazel~A Chieza, Krupa Prag, and Matthew Woolway.
\newblock An investigation of ibm quantum computing device performance on
  combinatorial optimisation problems.
\newblock \emph{arXiv preprint arXiv:2107.03638}, 2021.

\bibitem[{Blank} and {Deb}(2020)]{pymoo}
J.~{Blank} and K.~{Deb}.
\newblock pymoo: Multi-objective optimization in python.
\newblock \emph{IEEE Access}, 8:\penalty0 89497--89509, 2020.

\bibitem[Lam et~al.(2015)Lam, Pitrou, and Seibert]{lam2015numba}
Siu~Kwan Lam, Antoine Pitrou, and Stanley Seibert.
\newblock Numba: A llvm-based python jit compiler.
\newblock In \emph{Proceedings of the Second Workshop on the LLVM Compiler
  Infrastructure in HPC}, pages 1--6, 2015.

\bibitem[Akiba et~al.(2019)Akiba, Sano, Yanase, Ohta, and Koyama]{optuna_2019}
Takuya Akiba, Shotaro Sano, Toshihiko Yanase, Takeru Ohta, and Masanori Koyama.
\newblock Optuna: A next-generation hyperparameter optimization framework.
\newblock In \emph{Proceedings of the 25rd {ACM} {SIGKDD} International
  Conference on Knowledge Discovery and Data Mining}, 2019.

\bibitem[Coffey(2017)]{coffey2017adiabatic}
Mark~W Coffey.
\newblock Adiabatic quantum computing solution of the knapsack problem.
\newblock \emph{arXiv preprint arXiv:1701.05584}, 2017.

\bibitem[Burkard et~al.(1997)Burkard, Karisch, and Rendl]{burkard1997qaplib}
Rainer~E Burkard, Stefan~E Karisch, and Franz Rendl.
\newblock Qaplib--a quadratic assignment problem library.
\newblock \emph{Journal of Global optimization}, 10\penalty0 (4):\penalty0
  391--403, 1997.

\bibitem[Reinelt(1991)]{reinelt1991tsplib}
Gerhard Reinelt.
\newblock Tsplib—a traveling salesman problem library.
\newblock \emph{ORSA journal on computing}, 3\penalty0 (4):\penalty0 376--384,
  1991.

\bibitem[Verma and Lewis(2020)]{verma2020penalty}
Amit Verma and Mark Lewis.
\newblock Penalty and partitioning techniques to improve performance of qubo
  solvers.
\newblock \emph{Discrete Optimization}, page 100594, 2020.
\newblock ISSN 1572-5286.
\newblock \doi{https://doi.org/10.1016/j.disopt.2020.100594}.
\newblock URL
  \url{https://www.sciencedirect.com/science/article/pii/S1572528620300281}.

\end{thebibliography}

%\appendix
%\section{}

\end{document}